%% file: paper_arxiv.tex
\documentclass[runningheads]{llncs}
\usepackage{graphicx}

\usepackage{tikz}
\usepackage{comment}
\usepackage{amsmath,amssymb} %
\usepackage{color}
\usepackage{cite}
\usepackage{enumitem}
\usepackage{mathtools}
\usepackage{dsfont}
\usepackage{url}
\usepackage{microtype}
\usepackage[normalem]{ulem}

\usepackage[accsupp]{axessibility}  %

\definecolor{Yellow}{HTML}{fab802}
\definecolor{Red}{HTML}{c90404}
\definecolor{Green}{HTML}{0ead20}
\definecolor{Gray}{HTML}{484848}

\DeclareMathOperator*{\argmax}{arg\,max}

\usepackage{footmisc}
\DefineFNsymbols{mySymbols}{{\ensuremath\ast}{\ensuremath\dagger}
{\ensuremath\ddagger}{\ensuremath\mathsection}{\ensuremath\mathparagraph}
{\ensuremath\parallel}{**}{\ensuremath{\dagger\dagger}}{\ensuremath{\ddagger\ddagger}}}
\setfnsymbol{mySymbols}

\begin{document}
\pagestyle{headings}
\mainmatter
\def\ECCVSubNumber{4411}  %

\title{Fine-Grained Visual Entailment} %

\titlerunning{Fine-Grained Visual Entailment}
\author{Christopher Thomas\thanks{indicates equal contribution}\thanks{corresponding author} \and
Yipeng Zhang$^{\ast}$ \and Shih-Fu Chang}
\authorrunning{Thomas et al.}
\institute{Columbia University, New York, NY 10034, USA\\ 
\email{\{christopher.thomas, zhang.yipeng, sc250\}@columbia.edu}}
\maketitle
\vspace{-1.4em}
\input{abstract}
\input{intro.tex}

\input{related.tex}
\input{method.tex}

\input{results.tex}
\input{conclusions.tex}

\bibliographystyle{splncs04}
\bibliography{bibliography}

\end{document}


\pagestyle{headings}
\mainmatter
\def\ECCVSubNumber{4411}  %

\title{Fine-Grained Visual Entailment\\Supplementary Material} %

\titlerunning{Fine-Grained Visual Entailment}
\author{Christopher Thomas\thanks{indicates equal contribution}\thanks{corresponding author} \and
Yipeng Zhang$^{\ast}$ \and Shih-Fu Chang}
\authorrunning{Thomas et al.}
\institute{Columbia University, New York, NY 10034, USA\\ 
\email{\{christopher.thomas, zhang.yipeng, sc250\}@columbia.edu}}
\maketitle

\section{AMR Annotation and Statistics}

\subsection{Annotation details}

\begin{figure*}[th]
    \centering
    \includegraphics[width=1\textwidth]{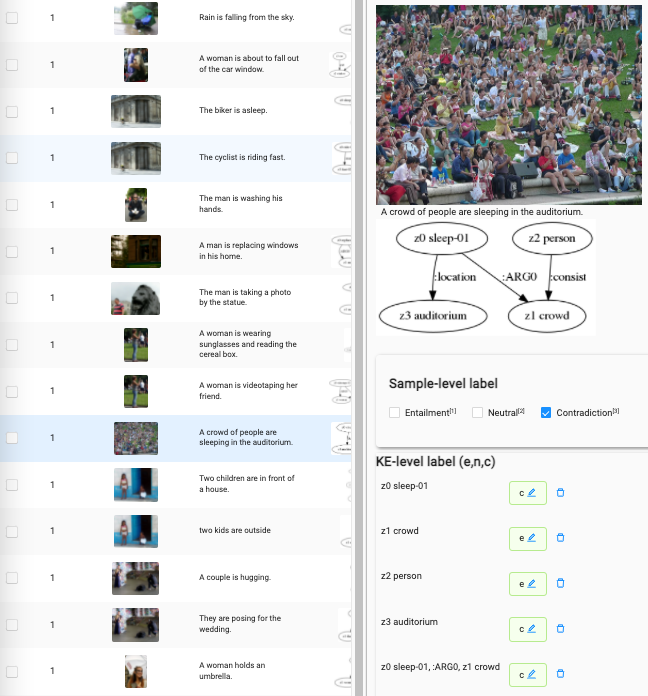}
    \caption{
    Screenshot of our annotation interface.
    }
    \label{fig:supp:annotation}
\end{figure*}

All our AMR strings used for annotation are produced by SPRING~\cite{bevilacqua-etal-2021-one}, which is a model that achieved recent SOTA performance on AMR semantic parsing, with a SMATCH score~\cite{cai2013smatch} of 83. %

Our annotation interface is shown in Figure \ref{fig:supp:annotation}. The image, the hypothesis, the AMR graph, and the extracted KEs are shown to the annotators. The annotators are required to follow the definitions in PropBank \cite{palmer2005proposition} and the AMR 3.0 specifications \cite{amr3} to annotate each KE (node or tuple), and provide the sample-level label. As mentioned in the main text, we provide the ``opt-out'' option for KEs. These opt-out KEs, in most cases, consist of nodes that require context (e.g., adjectives such as ``\texttt{big}'', numbers such as ``\texttt{2}'', time-indicators such as ``\texttt{about-to}''). These KEs are not considered during evaluation.
\FloatBarrier
\newpage

\subsection{Knowledge element distribution per category}
\begin{figure*}[th]
    \centering
    \begin{subfigure}[t]{0.55\textwidth}
        \includegraphics[width=\linewidth]{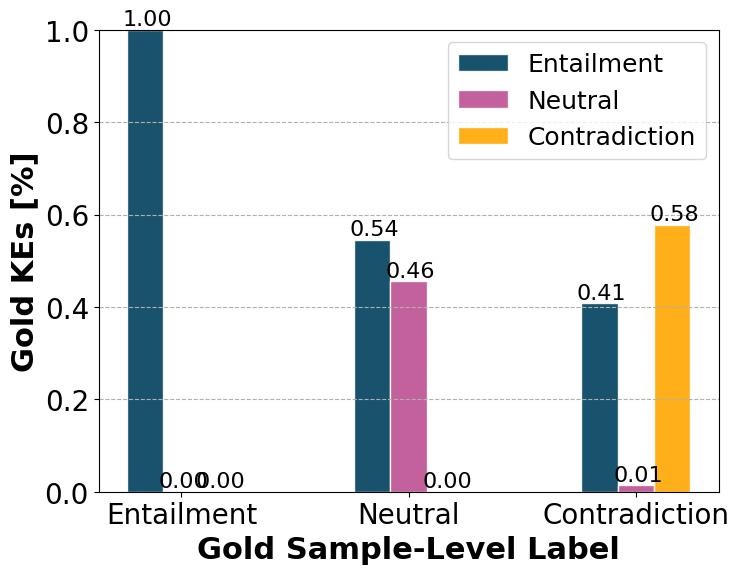}
        \caption{KE labels.}
        \label{fig:supp:dist:gold}
    \end{subfigure} %
    \centering
    \begin{subfigure}[t]{0.55\textwidth}
        \includegraphics[width=\linewidth]{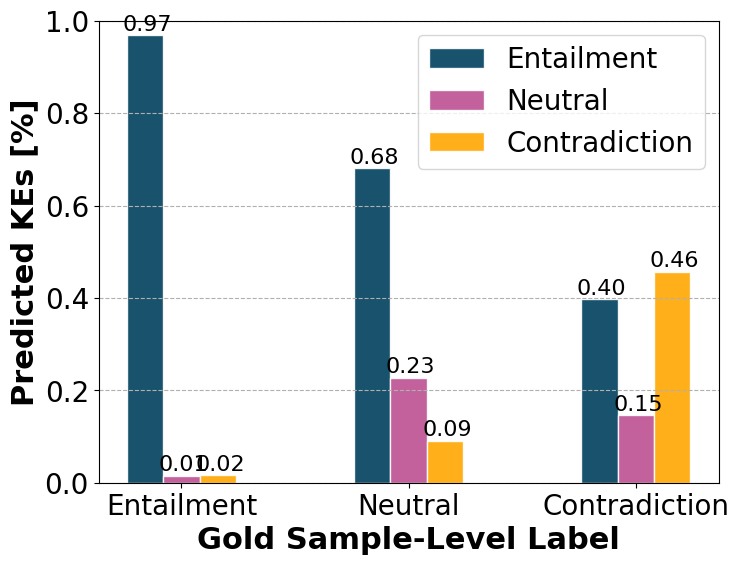}
        \caption{KE predictions by our model.}
        \label{fig:supp:dist:ours}
    \end{subfigure}
    \centering
    \begin{subfigure}[t]{0.55\textwidth}
        \includegraphics[width=\linewidth]{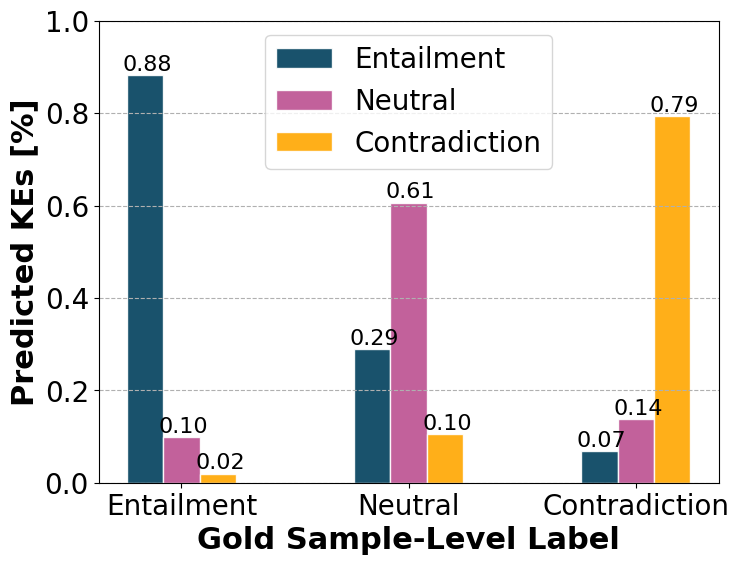}
        \caption{KE predictions by VE+AMR$\to$KE.}
        \label{fig:supp:dist:baseline}
    \end{subfigure} %
    \caption{Distributions of (a) ground truth KE labels, (b) predicted KEs by our model within each sample category, and (c) predicted KEs by the VE+AMR$\to$KE model within each sample category.}
    \label{fig:supp:dist}
\end{figure*}

In Figure \ref{fig:supp:dist}, we show a breakdown of the distributions of both the knowledge element level annotations and the predictions by our method and the most competitive baseline on the KE-level. We show the normalized breakdown for each sample-level label (on the x-axis). For example, the three bars above the ``contradiction'' label on the x-axis indicate the distribution of KEs for those samples labeled contradiction. That is to say, of all the KEs in samples labeled contradiction, what percentage were labeled entailment (blue), neutral (purple), or contradiction (yellow).

In Figure \ref{fig:supp:dist:gold}, we show the distribution of human annotated KEs (gold KEs) for each gold sample-level label. Note that the distribution for each sample-level label adds to one and that our gold labels exhibit no MIL violations (e.g.,~no neutral or contradiction KEs in samples labeled entailment). Significantly, we observe that entailed KEs make up a significant portion of both ``neutral'' and ``contradiction'' samples. For samples labeled neutral, there are more KEs labeled entailment than there are KEs labeled neutral (54\% versus 46\%). Similarly, for contradiction KEs, 58\% of KEs are labeled contradiction and 41\% are labeled entailed. If a method was perfectly accurate at predicting the sample-level label, one would get 54\% of KEs wrong for neutral samples and 42\% of KEs wrong for contradiction samples. This underscores the importance of our method of making fine-grained, KE-level predictions. 

One surprising observation in Figure \ref{fig:supp:dist:gold} is that there are very few ``neutral'' KEs in samples labeled contradiction. By definition, contradiction samples can contain entailed KEs (true claims about the image), neutral KEs (claims that could be true), and should contain at least one contradiction KE (a false claim). When Turkers were tasked with creating the SNLI dataset, they were prompted to ``Write one alternate caption that is definitely a false description of the photo'' for contradiction \cite{bowman2015large}. To do so, we observed that Turkers would often mention something specific truly in the the image (i.e.,~that was entailed), but then make an obviously false claim about it. Turkers usually didn't elaborate by adding additional ``neutral'' information into their hypotheses. However, because we neutral information \textit{can} technically appear in a contradiction sample, we still allow our model to predict neutral KEs for contradiction samples. This lack of neutral KEs in contradiction samples is dataset specific and likely due to the way the dataset was constructed.

In Figure \ref{fig:supp:dist:ours}, we show our method's KE-specific distributions across the three gold sample-level labels. We observe our distribution is close to the gold distribution (Figure \ref{fig:supp:dist:gold}), with a few differences. For neutral, our method is overconfident about entailed KEs, predicting 68\% of KEs as entailed, compared to the 54\% which are actually entailed. In practice, our method confuses a number of truly neutral KEs as entailed. Note distinguishing some neutral KEs from entailed is often challenging because it requires one to distinguish whether a KE is true or merely could be true which is sometimes subjective. We observe that our method mistakenly predicts 9\% of KEs in the neutral class as contradiction. For contradiction, we observe our method predicts 40\% of KEs as entailed which closely tracks the amount in the gold set (41\%). We observe our model predicts 15\% of KEs as neutral in contradiction. As stated above, we note that we could significantly improve our model's performance further on this dataset by enforcing a constraint that no neutral KEs were allowed in contradiction samples, but this would be a dataset-specific constraint. We instead chose to impose constraints consistent with the logical definitions that define the fine-grained visual entailment problem rather than tailor them to a particular dataset.  Had these incorrectly predicted neutral KEs instead been predicted contradiction, our distribution for contradiction would be 61\%, close to the 58\% in the ground truth distribution.

In Figure \ref{fig:supp:dist:baseline}, we show the KE distribution of our most competitive baseline on the KE-level (VE+AMR$\to$KE). We observe a significant degradation in performance compared to the gold distribution. For the entailment category, 10\% of KEs are mistakenly predicted neutral. For the neutral category, the model predicts only 29\% of KEs as entailed (54\% in gold), while predicting 61\% as neutral (46\% in gold). This problem is most acute for the contradiction category where the model predicts 79\% of KEs as contradiction (58\% in gold) and only 7\% as entailed (41\% in gold). Without our KE-level constraints, the model's KE-level predictions are much more frequently the same as the the sample-level label than compared to our method or the gold labels, suggesting the predictions are not as semantically meaningful. 

These distributions also explain why the baselines \textit{appear} misleadingly strong at accurately predicting neutral and contradiction in Table 1 in the main text (left group of results), but then perform much worse overall. Because contradiction KEs only appear in contradiction samples and the model predicts the vast majority of KEs in contradiction samples as contradiction, the accuracy of the model on the set of contradiction KEs is very high (i.e., due to a high recall). However, the model is actually performing very poorly overall on contradiction samples (shown by the distribution), infrequently predicting any entailed KEs, when in actuality 41\% of KEs are entailed. The same reasoning equally applies to the neutral case, where the baseline only predicts 29\% of KEs as entailed in neutral samples, when in actuality 54\% should be entailed.

\FloatBarrier
\newpage
\input{_tab_ablation}

\section{Ablation Study}
We show ablation of different loss weights for our three losses in Table~\ref{table:supp:ablation}. First, competition exists between $\mathcal{L}_{\text{CLS}}$ and the other two losses, so we have to weigh $\mathcal{L}_{\text{CLS}}$ down for the model to achieve high KE performance. In other words, our model must trade off between focusing on the sample-level task and the KE-level task. This, however, does not undermine our performance on the sample level -- setting $\beta_{\text{CLS}}$ to 0.5 gives us a sample-level performance of 80.73\% using only the KE predictions, best among all the methods studied in this paper. We also find that setting $\beta_{\text{CLS}}$ too low (0.1) hurts the performance on the neutral and contradiction samples, as well as on the sample level (compared to $\beta_{\text{CLS}}=0.5$). In general, these two models show comparable performance. 

Structural constraints are essential for our model. If we remove the structural constraint, the model shows very low performance on both the neutral and contradiction samples; its structural accuracy drops to 70.26\%, which means that the predictions might not be meaningful. On the other hand, putting less weight on $\mathcal{L}_{\text{CLS}}$ (0.5) and $\mathcal{L}_{\text{KE}}$ (0.5) and keeping $\beta_{\text{STRUC}}=1$ yields lower performance on the KEs (61.68\%), albeit reaching very high structural accuracy (98.4\%). In summary, we find it beneficial to keep an equal weight on $\mathcal{L}_{\text{KE}}$ and $\mathcal{L}_{\text{STRUC}}$ and weighing $\mathcal{L}_{\text{CLS}}$ slightly lower.

\section{Additional Implementation Details}

\subsection{AMR tokenization}
The AMR string produced by SPRING~\cite{bevilacqua-etal-2021-one} is separated by the new line character, ordered according to depth-first search (DFS). We use the SPRING model pretrained on AMR 3.0 and the \texttt{amrlib} library\footnote{https://github.com/bjascob/amrlib} for all AMR extractions. 

We then remove the newline characters and the redundant token ``\textbf{/}''. Next, we merge the roles ``\texttt{:opx}'' into ``\texttt{:op}'' (\texttt{x} is an index) and roles ``\texttt{:sntx}'' into ``\texttt{:snt}'' because these indices do not affect the meaning conveyed. For example, ``\texttt{:op1}'' and ``\texttt{:op2}'' can be used interchangeably; on the other hand, ``\texttt{:ARG0}'' usually means the subject of a verb while ``\texttt{:ARG1}'' usually means the object, so the indices on ``\texttt{:ARGx}'' are essential for semantic representation. Note that the predicate labels (e.g., ``\texttt{z0}'') are also important because they provide information of cross-referencing.

Finally, We remove the predicate (node) indices (e.g., ``\texttt{-01}'' in ``\texttt{walk-01}'') used for disambiguating different meanings for predicates. This is because we rely on the transformer for disambiguation, given their capability of disambiguating different meanings for words in the English corpus. We also find such approach work better empirically. 

These changes not only shorten the AMR sequences, but also create more instances of ``\texttt{:op}'', ``\texttt{:snt}'', and the predicate tokens, which helps learning. We add all the edge tokens, the ``\texttt{amr-unknown}'' predicate, as well as the role-inversion indicator ``\texttt{-of}'' into the vocabulary and initialize each newly added token's embedding as the average embedding of their word pieces (pretrained by Oscar+). We do not add new tokens for predicates other than ``\texttt{amr-unknown}''.

\subsection{Other implementation details}

Oscar+ is pretrained on three sequences (object features, tags, text), with the tag sequence and text sequence using distinct token type embeddings. When we add another sequence, the AMR sequence, we create a new token type embedding and initialize it with the pretrained token type embedding of the text. All the parameters in the model are finetuned on our task.

We follow Oscar+~\cite{zhang2021vinvl} and uses object features of dimension 2054. Among it, 2048 are the extracted region features; four dimensions are the bounding box coordinates (top, left, bottom, right) normalized by image size; the rest two dimensions are the normalized height and width of the object. The object detector's confidence threshold is set to 0.2, as in Oscar+~\cite{zhang2021vinvl}.

Our losses are computed on KEs. However, in rare cases when the sequence is too long, some ending KE tokens could get truncated. If a sample with neutral or contradiction label has such tokens, we do not enforce KE or structural losses on them so as not to introduce false signals.

Before summing up the losses, we take the average of $\mathcal{L}_{\text{CLS}}$ and $\mathcal{L}_{\text{KE}}$ across each sample and $\mathcal{L}_{\text{STRUC}}$ across each relation. Since we do not have fine-grained KE labels for the training set, the checkpoint that performs the best on the sample level using $f_{\text{CLS}}$ (on the validation set) is saved for evaluation.

\newpage
\section{Additional Qualitative Results}

\begin{figure*}[ht]
    \centering
    \includegraphics[width=1\textwidth]{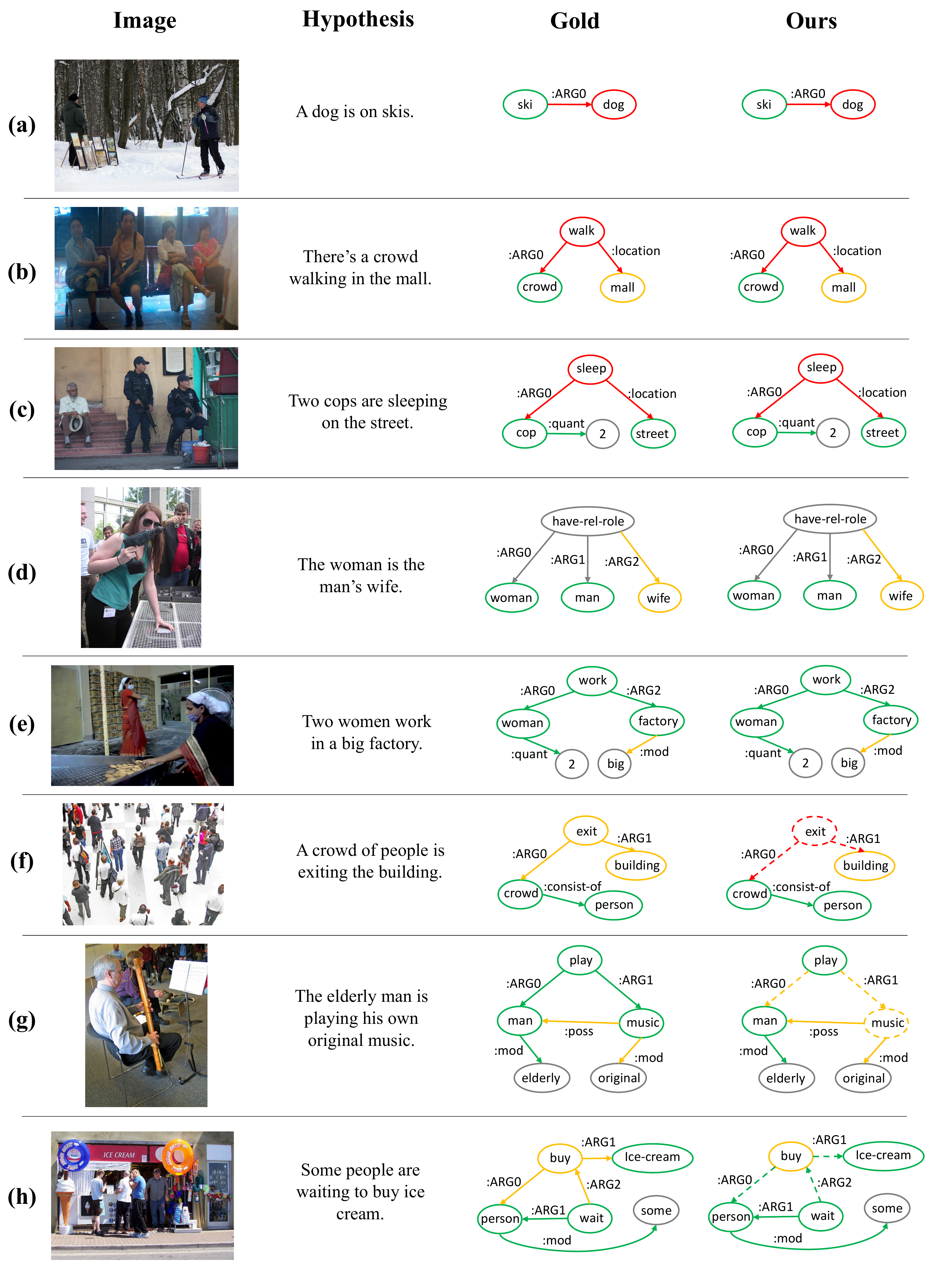}
    \caption{
    Additional qualitative results showing our KE-level predictions on AMR graphs. Nodes and edges (representing tuples) are colored based on their predicted label ({\color{Green} ent}, {\color{Yellow} neu}, {\color{Red} con}, {\color{Gray} opt-out}). Wrong predictions are denoted by dashed lines. We drop predicate labels (e.g., \texttt{z0}) for brevity. 
    }
    \label{fig:supp:qual}
\end{figure*}

We show additional qualitative results in Figure \ref{fig:supp:qual}. As in our main text, the color of the edges (representing the tuple) and nodes indicate the predicted label. A dotted line indicates an incorrectly predicted KE. We observe that our model is able to accurately identify the source of inconsistencies and logically reason about the relationship of the image to each hypothesis.

In Figure \ref{fig:supp:qual} (a), we observe the model identifies that there is no dog and that no dog is skiing. In (b), the model identifies that no one is walking, is unsure about the location, but does detect a crowd of people. In (c), the model identifies no one is sleeping, but correctly determines there are two police officers and there is a street. In (d), several tuples and the have-rel-role node can't be predicted without additional context, but the model accurately determines that the wife relationship is unknown. In (e), our model correctly concludes the action (work) is entailed, that there are two women, and they are working in a factory. Note that the model is neutral as to whether the factory is big, as is the human annotator. In (f), we see our model mistakenly concludes that the people are not ``exiting.'' We see that some individuals are standing or staring, but whether they are actually exiting is unclear. The model's understanding of exiting may be that exiting only occurs when someone is walking out a door, for example, and thus the model predicted contradiction. In (g), we see a another mistake of our method. We observe that our model believes someone is ``playing'', but is unclear if it is the man who is playing. The model is unclear as to whether music is present or if it is being played. However, the model correctly identifies that the man is elderly and can't determine if the music is original. The model may have become confused in this example by the unusual instrument and the position of objects. Lastly, in (h) we observe one of the rare structural violations from our model. The model is unclear as to whether ``buying'' is occurring, but predicts ``person buying'' and ``waiting to buy'' as entailed, resulting in a bottom-up structural violation.

\FloatBarrier
\newpage
\bibliographystyle{splncs04}
\bibliography{bibliography}

%% file: abstract.tex
\begin{abstract}
Visual entailment is a recently proposed multimodal reasoning task where the goal is to predict the logical relationship of a piece of text to an image. In this paper, we propose an extension of this task, where the goal is to predict the logical relationship of fine-grained knowledge elements within a piece of text to an image. Unlike prior work, our method is inherently explainable and makes logical predictions at different levels of granularity. Because we lack fine-grained labels to train our method, we propose a novel multi-instance learning approach which learns a fine-grained labeling using only sample-level supervision. We also impose novel semantic structural constraints which ensure that fine-grained predictions are internally semantically consistent. We evaluate our method on a new dataset of manually annotated knowledge elements and show that our method achieves 68.18\% accuracy at this challenging task while significantly outperforming several strong baselines. Finally, we present extensive qualitative results illustrating our method's predictions and the visual evidence our method relied on. Our code and annotated dataset can be found here: \url{https://github.com/SkrighYZ/FGVE}.
\end{abstract}

%% file: intro.tex
\section{Introduction}
\label{sec:intro}

Tasks requiring multimodal understanding across vision and language have seen an explosion of interest in recent years, driven largely by their many downstream applications. Common tasks  include visual question answering \cite{antol2015vqa, gokhale2020vqa, sheng2021human}, visual commonsense reasoning \cite{zellers2019recognition}, and visual dialog \cite{das2017learning, Das_2017_CVPR}. Moreover, tasks that had historically been studied by only the natural language processing or computer vision communities have recently received attention from both communities. For example, event extraction \cite{zhang2017improving, li2020cross, chen2021joint} and coreferencing \cite{kong2014you, plummer2017phrase}, longstanding information extraction and NLP tasks, have all recently been explored multimodally.

\begin{figure*}[t]
    \centering
    \includegraphics[width=0.9\textwidth]{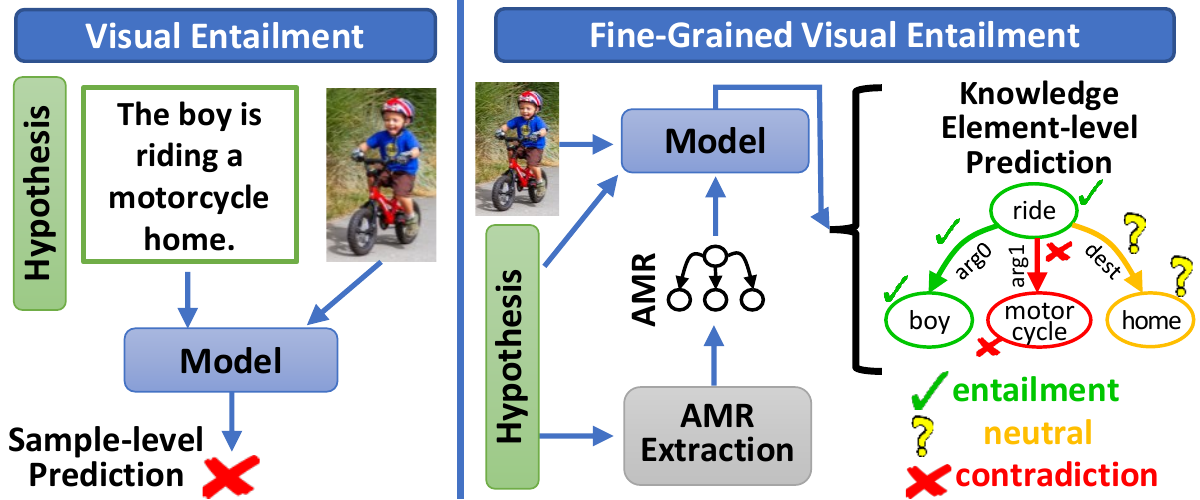}
    \caption{
    In the standard visual entailment task, a model predicts the logical relationship of the entire hypothesis to an image. In our proposed task, the model predicts the relationship of each knowledge element of the hypothesis. Specifically, our model predicts a boy is present, someone is riding, the boy is riding, a motorcycle is not present, no one is riding a motorcycle, and can't conclude whether someone is riding home or not.
    }
    \label{fig:intro:concept}
\end{figure*}

One such task is the textual entailment task, first proposed in 2005 \cite{dagan2005pascal}. The task requires a system to decide whether a piece of text (the hypothesis) can be logically deduced from another piece of text accepted as true (the premise). 
Xie et al.~\cite{xie2019visual} posed a multimodal variant of the task called visual entailment, which replaces the textual premise with an image. Because of the rich cross-modal reasoning required, it has become a standard benchmark for testing joint vision-and-language understanding in many recent multimodal models \cite{huang2021seeing, chen2020uniter, lu202012}.

The standard visual entailment benchmark \cite{xie2019visual} is a three-way classification task between entailment (hypothesis is true), neutral (hypothesis could be true or false), and contradiction (hypothesis is false). Though the classification is made for the entire hypothesis, the task by its very nature requires fine-grained multimodal reasoning. For example, for a hypothesis to be labeled contradiction, the model must find at least one facet of the hypothesis that conflicts with the image. 
The task as posed, however does not require the model to produce fine-grained predictions which would explain its reasoning. 

This lack of fine-grained predictions is significant for several important reasons. First, it limits the utility of the task to downstream applications. Fundamentally, the visual entailment task requires models to search for visual evidence necessary to make logical inferences about the text. This capability has many possible downstream use cases, from detecting image-text inconsistencies for misinformation detection \cite{luo2021newsclippings} to ensuring answers to questions are entailed by the image \cite{si2021check}. However, many of these tasks require localized predictions \cite{fung2021infosurgeon}, which existing methods are unable to provide. Secondly, the fine-grained predictions produced by the model naturally explain its reasoning, making 
its prediction interpretable.

To address the above shortcomings, we propose the \textbf{fine-grained visual entailment} task. Similar to the original visual entailment task, our goal is to predict the logical relationship of textual claims about images. But 
differently, we require the model to make predictions for each specific ``claim'' in the text. We illustrate our task in Figure \ref{fig:intro:concept}.
Our method works by decomposing the textual hypothesis into its constituent parts which we call ``knowledge elements'' (KEs). Knowledge elements are the claims that collectively constitute the entire hypothesis' meaning. In order to decompose the hypothesis into its constituent KEs, we represent the hypothesis as its abstract meaning representation (AMR) \cite{banarescu2013abstract, amr3} graph. We choose AMR to represent the semantics of the hypothesis because AMR captures the semantic meaning of text irrespective of its syntax %
\cite{banarescu2013abstract}. 

We train a multimodal transformer to make fine-grained predictions on nodes and tuples (i.e.~the KEs) within the AMR graph. Our transformer takes as input visual tokens, the hypothesis, and a linearized representation \cite{matthiessen1991text, goodman2020penman} of the AMR graph. 
To make predictions for each KE within the AMR graph, we introduce a novel local aggregation mechanism which learns a localized contextual representation. 
Our contextual representation fuses the representation of the KE's AMR with its associated visual context. These representations are then used to make KE-specific predictions.

The model described above requires KE-level supervision to train, but our dataset only contains \textit{sample}-level supervision.
Rather than rely on expensive and hard to obtain AMR graph annotations, we instead propose a novel multi-instance learning (MIL) approach. Specifically, we leverage the sample-level label to impose a set of MIL constraints on our prediction function which induce a fine-grained labeling of the graph without requiring any new annotations.

While our MIL losses ensure that knowledge element-level predictions are consistent with the sample-level label, they do not ensure that they are semantically consistent with each other. 
Thus, we impose both top-down and bottom-up semantic structural constraints which penalize the model for semantically inconsistent knowledge element predictions. These constraints leverage the same intuition as our sample-level MIL constraints, but instead work \textit{between} KEs %
at different structures within the AMR graph.

In order to benchmark performance on this task, we densely annotate AMR graphs at the knowledge element-level. We compare our method against a number of baselines across numerous metrics. Experiments show that our approach substantially outperforms all baselines for the fine-grained visual entailment task. We also include detailed qualitative results showing our method produces semantically plausible predictions at the knowledge element-level as well as examples of the ``visual evidence'' chosen by our model to make its predictions.

The major contributions of this paper are as follows:
\begin{itemize}[nolistsep,noitemsep]
\item We introduce and formulate the novel task of fine-grained visual entailment. Our goal is to make fine-grained judgments about the logical relationship of an image to each knowledge element within a piece of text. 
\item We contribute a fine-grained visual entailment benchmark of AMR graphs densely annotated by experts 
to facilitate research on this new task.
\item We propose a novel method for this task which relies on localized cross-modal contextual representations of each knowledge element. 
\item We develop a number of novel loss functions to train our method to make knowledge element-level predictions with only sample-level labels. We also devise novel structural constraints which ensure that our model's predictions are internally semantically consistent.
\item We perform a detailed experimental evaluation of our method with a number of baselines which clearly demonstrates the superiority of our approach. We also perform ablations of different aspects of our model and loss functions. 
\item Finally, we present qualitative examples showcasing our results and provide examples of the ``visual evidence'' used by our method to make its predictions.
\end{itemize}

%% file: related.tex
\section{Related work}
\label{sec:rel_work}

\noindent \textbf{Textual entailment.}
Textual entailment (predicting whether a hypothesis is entailed by a premise) has long been studied by the natural language processing \cite{dagan2005pascal} community. 
Later work such as SICK \cite{marelli2014sick} and the Stanford natural language inference benchmark (SNLI) \cite{bowman2015large} expanded the task definition to allow more granular labels
, i.e.~by adding the neutral category. 
The SNLI benchmark is a large-scale benchmark of crowdsourced hypotheses written for Flickr30K \cite{young2014image} image captions (which served as the premises). 
\cite{camburu2018snli} further extend the SNLI dataset with human-written free-form text explanations of the sample's label. None of these works operate multimodally or make granular predictions as we do.

\noindent \textbf{Visual entailment.}
Most related to this paper is past work in visual entailment. Xie et al.~\cite{xie2019visual} introduced the visual entailment task which replaced the textual premise from SNLI (a Flickr30K image caption) with its corresponding image, while preserving the original label. Other work \cite{chen2021explainable, liu2020violin} has also explored visual entailment in the video domain. \cite{kayser2021vil} observed that the visual entailment dataset contained substantial label errors caused by replacing the image caption with its image. To correct this, \cite{kayser2021vil} reannotated the samples labeled neutral in the test set and proposed an automatic technique to correct some mislabeled neutral train examples. 
\cite{kayser2021vil} also use the rationales from \cite{camburu2018snli} to train a text generation method to generate free-form textual ``explanations'' of their predictions. 

Our approach offers several significant benefits over \cite{kayser2021vil}'s. First, \cite{kayser2021vil} train a text generator supervised by human-written rationales and conditioned on their visual entailment model's embeddings. While this generates natural language explanations, there is no guarantee that the generated text truly describes the model's reasoning.  
Moreover, the generated text may not address specific claims (or any claims) made in the hypothesis. Unlike prior work, our method decomposes the hypothesis into its constituent KEs. Our KE-level predictions naturally cover all claims within the hypothesis which may be important for downstream applications, while inherently explaining the model's reasoning.

\noindent \textbf{Explainability.}
Our work is also related to research in producing explainable and interpretable predictions. 
Common examples include as saliency-map techniques \cite{shrikumar2017learning, zhou2018interpreting, zunino2021explainable} as well attention mechanism visualizations \cite{fukui2019attention, li2018occlusion}. 
One popular such example of the former category is Grad-CAM \cite{selvaraju2017grad} which computes class-specific gradient heatmaps with the input. \cite{wagner2019interpretable} produce fine-grained visual explanations of image regions which caused the model to predict a particular class. 
More recent work visualizes the attention maps in transformers \cite{chefer2021generic, voita2019analyzing, chefer2021transformer}. Similar to our method, \cite{chen2021explainable} produce grounded video regions as explanations for video entailment, but do not tackle the fine-grained entailment setting as we do. \cite{fung2021infosurgeon} make fine-grained predictions of image-text inconsistency using a predefined ontology, but do not consider the open domain and more granular entailment problem we do.

\noindent \textbf{Multi-instance learning.}
Our model is required to learn which knowledge elements are entailed, neutral, or contradictory from only the sample-level label. Our work is thus related to multi-instance learning (MIL) methods where a bag of samples are assigned a single label with at least one sample in the bag being the label of the bag \cite{dietterich1997solving}. MIL methods have recently been explored for a variety of tasks including image classification \cite{rymarczyk2021kernel, wu2015deep}, object detection \cite{yuan2021multiple, yang2019detecting, dong2021boosting}, scene graph generation \cite{shi2021simple}, and video segment localization \cite{zhou2017adaptive, luo2020weakly}. All share the goal of learning a finer-grained prediction function than directly available from the training labels. We are the first to apply MIL techniques to learn a knowledge element-level entailment prediction model. 

%% file: method.tex
\section{Fine-grained Visual Entailment}
\label{sec:method}
Given an image and a textual hypothesis about the image, our goal is to predict the logical relationship of the image to every ``assertion'' contained within the hypothesis. To do so, we transform the textual hypothesis into its abstract meaning representation (AMR) graph. We make predictions for each node and tuple (edge with its endpoint nodes) in the graph, which we call knowledge elements (KEs). KEs consist of assertions within the hypothesis about actions, entities (objects, people, etc.), colors, gender, count, etc. We propose a multimodal transformer which operates over the image, hypothesis, and a representation of the AMR graph. At a high level, our method works by locating KE tokens in the AMR graph, aggregating visual information to create a contextualized KE embedding, and performing predictions by a classifier trained with novel multi-instance learning and structural losses (which enforce semantic consistency). 

\subsection{Problem formulation}
More formally, let $\mathcal{D} = \left\{\left(i^{(1)}, h^{(1)}, y^{(1)}\right), \ldots, \left(i^{(s)}, h^{(s)}, y^{(s)}\right), \ldots, \left(i^{(n)}, h^{(n)}, y^{(n)}\right) \right\}$ represent a dataset of image, hypothesis, and entailment label triples respectively, where $n$ is the number of sample triples within the dataset. The goal of the standard visual entailment task \cite{xie2019visual} is to learn the prediction function $f_\theta(i^{(s)}, h^{(s)}) = y^{(s)}$, i.e.,~to predict the sample-level logical relationship of the image and hypothesis. 
Note that in the remainder of this text, we omit the sample index and refer to a single sample for clarity unless noted.
Because we seek to make sub-hypothesis-level predictions, we first decompose each hypothesis into its constituent KEs. We denote the set of KEs extracted from $h$ as $KE = \left\{ ke_j \right\}_{j=0}^{|KE|}$. We seek a prediction function $g_\theta \left(i, h \right) = \left\{ y_{ke_{j}} \right\}_{j=0}^{|KE|}$ where $y_{ke_j}$ is the label of $ke_j$ describing its specific logical relationship with the image.

\begin{figure*}[t]
    \centering
    \includegraphics[width=1\textwidth]{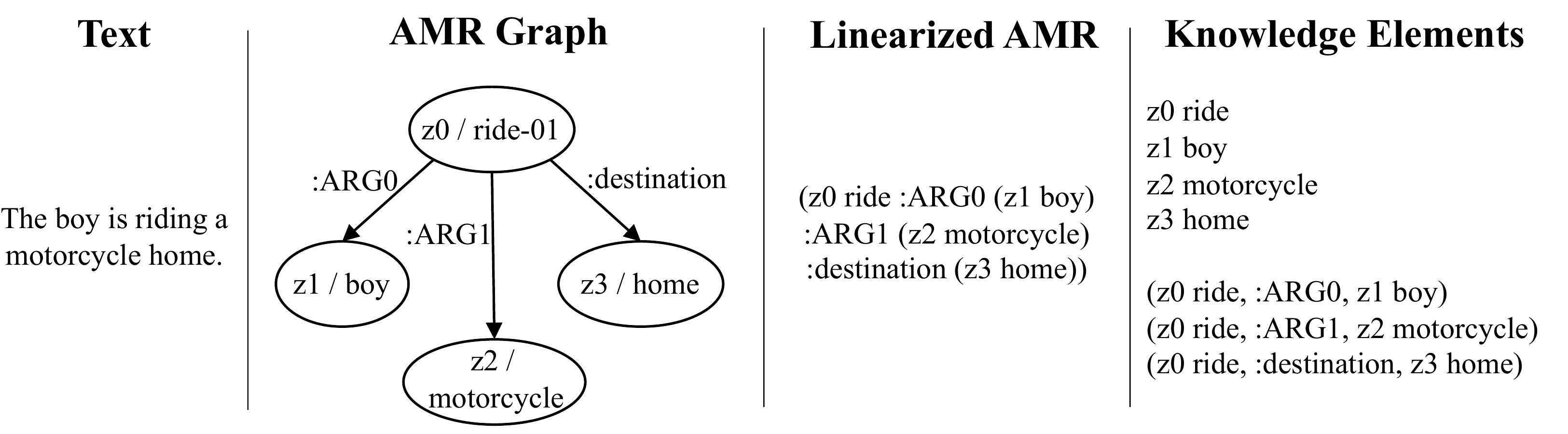}
    \caption{
    Illustration of knowledge element (KE) extraction. During preprocessing, we make simplifications to the linearized AMRs such as removing ``/'' and ``-01''. We include details on preprocessing and tokenization in our supplementary.
    }
    \label{fig:method:amr}
\end{figure*}

\subsection{Knowledge element extraction}
We next describe more specifically how we extract KEs from the hypothesis. Let $\mathcal{G}$ be a text to AMR graph prediction method, then $\mathcal{G}\left(h\right) = G$ represents the conversion of a hypothesis into its AMR graph representation $G$, where $G = (V,E)$ and $V$ and $E$ represent the set of vertices (nodes) and edges within the graph, respectively. Each node $v \in V$ represents a simple, atomic statement about the image (e.g.,~``there is a car''; ``something is walking''). In contrast, each edge $\overrightarrow{e} \in E$ is directed and defines the relationship type between nodes. Because edges are of ambiguous meaning without the nodes they connect, we do not consider edges independently as KEs. Instead, we consider the set of directed node-edge tuples denoted $T = \left( v_h, e, v_t \right)$ with $\left\{v_h,v_t\right\} \in V$, $v_h$ and $v_t$ being the head and tail nodes defined by the edge's direction, and $\overrightarrow{e}$ defining the relation type between the nodes. Each tuple thus represents a composite statement %
(e.g.,~``$v_t$ is performing the action in $v_h$''; ``$v_t$ is the color of $v_h$'').
Thus, the set of KEs extracted for a sample is $KE = \left\{ V \cup T \right\} = \left\{ ke_j \right\}_{j=0}^{|KE|}$. We show an example of what our AMR to KE extraction process looks like in Figure \ref{fig:method:amr}.

\subsection{Architecture}
\begin{figure*}[t]
    \centering
    \includegraphics[width=1\textwidth]{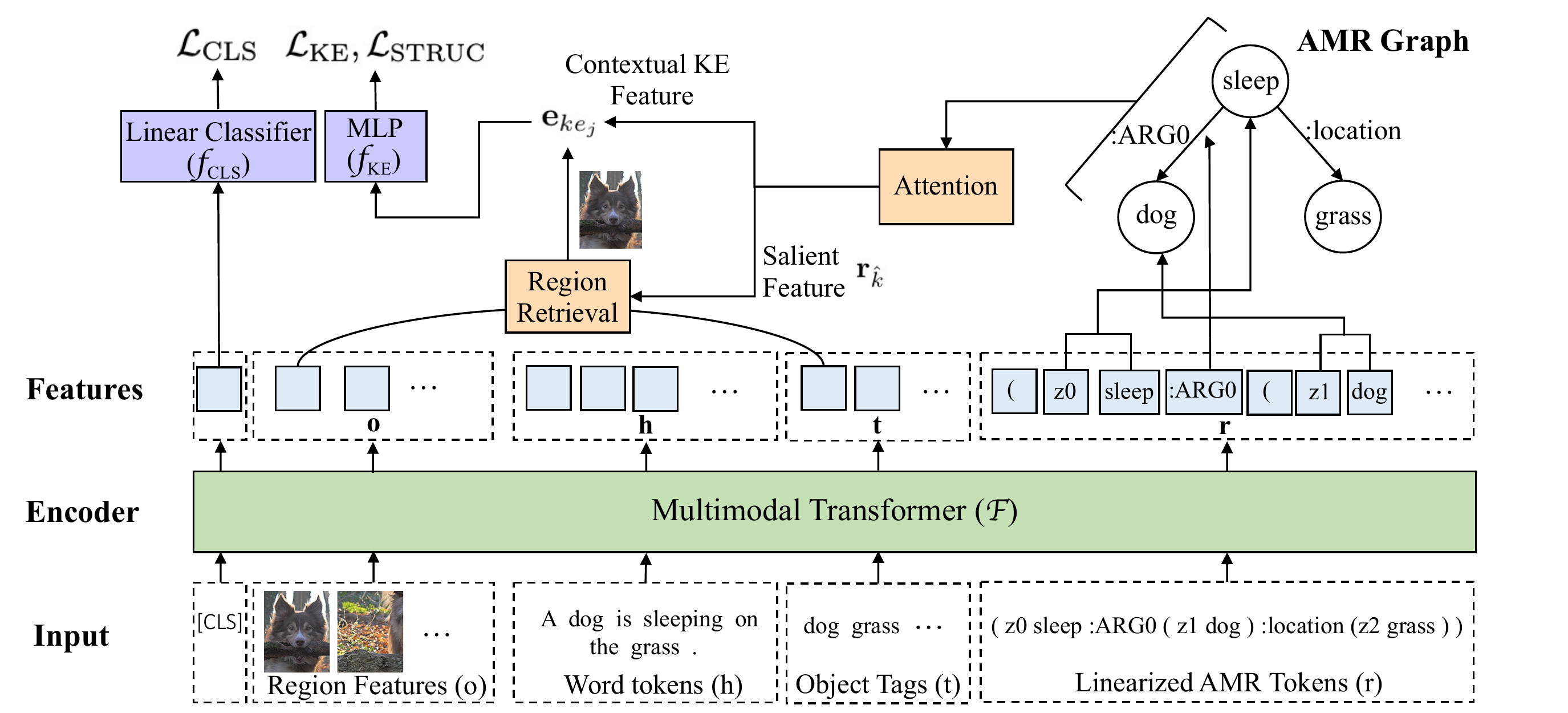}
    \caption{
    Our method that makes predictions on both the sample level and the KE level. We show the processing steps for an example KE: \texttt{(z0 sleep, :ARG0, z1 dog).}
    }
    \label{fig:method:architecture}
\end{figure*}

Figure~\ref{fig:method:architecture} shows the architecture of our method. In this work, we use OSCAR$^+$ \cite{zhang2021vinvl} as our multimodal encoder $\mathcal{F}$. OSCAR$^+$ achieved recent SOTA performance on several downstream vision-language tasks \cite{zhang2021vinvl}.
Given an image-hypothesis pair $(i, h)$, a pretrained object detector first extracts a sequence of object region features $o = o_1, \ldots, o_m$ and a sequence of predicted object tags $t = t_1, \ldots, t_p$ (object labels in text form) from the image, where $m$ and $p$ are the number of regions and tags respectively. Let $f_\varphi$ denote a graph linearization method \cite{matthiessen1991text}, then $r = f_\varphi(\mathcal{G}(h))$
where $r$ is $h$'s AMR in linearized form. The linearized AMR encapsulates all KEs within the hypothesis' AMR in a string form, while retaining their semantic structure.
We extract the token embeddings from the last layer of the encoder for each of the sequences: $(\mathbf{o}, \mathbf{t}, \mathbf{r}, \mathbf{h}) = \mathcal{F}(o, t, r, h)$. Although providing $h$ to the model is not required, it provides context and we find it slightly improves sample-level performance ($\sim 2\%$) in practice. Our method for fine-grained KE prediction to be described does not involve or require $\mathbf{h}$.

\subsection{Knowledge element-contextual aggregation}
Although multimodal interactions happen naturally through self-attention in the transformer's layers, we find it beneficial to apply attention to tokens inside each KE. For example, the model might pay more attention to predicates than entity labels such as ``\texttt{z0}''. 
Let $\mathbf{r} = (\mathbf{r}_1, \mathbf{r}_2, \ldots, \mathbf{r}_l)$ be the full AMR embedding sequence, where $l = |r|$. 
Let a subset $\mathbf{r}_{ke_j} = (\mathbf{r}_{l_1}, \ldots, \mathbf{r}_{l_k}) \in \mathbb{R}^{m_j \times d}$, where $d$ is the hidden state dimension and $\{l_1, \ldots, l_k\} \subseteq \{1, \ldots, l\}$, be the embeddings of the $m_j$ tokens (not necessarily consecutive) that form $ke_j \in KE$. 
To estimate each token's importance, we learn a function $f_{\alpha}$ that takes as input $\mathbf{r}_{ke_j}$ and outputs token-wise attention weights $\mathbf{w}_j \in R^{m_j}$. 
In this work, $f_{\alpha}(\mathbf{r}_{ke_j}) = \sigma(\mathbf{r}_{ke_j}\mathbf{w}^\alpha)$ , where $\sigma$ is the softmax function and $\mathbf{w}^\alpha \in \mathbb{R}^d$ is a learned vector shared by all KEs. We leave exploring stronger weighting mechanisms as future work.

A challenging aspect of our task requires the model to create grounded representations for individual KEs. For a single node in the AMR graph, context information from the image space could be essential for prediction. Therefore, we propose a method to retrieve relevant image region features for each $\mathbf{r}_{ke_j}$. 

We first select the most salient token $\hat{k}$ characterized by having the highest attention weight ($\hat{k} = \argmax_k w_{j, k}$, where $w_{j, k}$ is the $k$-th element of $\mathbf{w}_j$). We then compare the cosine similarity between $\mathbf{r}_{\hat{k}}$ and each of the tag embeddings $\mathbf{t}_{k'}$, $\frac{\mathbf{r}_{\hat{k}} \cdot \mathbf{t}_{k'}}{\| \mathbf{r}_{\hat{k}}\|\|\mathbf{t}_{k'}\|}$, and retrieve the most similar tag $t_{\hat{k}'}$. With the correspondence given by the object detector, we can now retrieve the object region feature $\mathbf{o}_{\hat{k}''} \in \mathbb{R}^d$ that $t_{\hat{k}'}$ refers to. 

The full contextualized embedding used for predicting $ke_j$'s label is therefore obtained by:
\begin{equation}
\mathbf{e}_{ke_j} = \text{Concatenate}([\mathbf{r}_{ke_j}^\top\mathbf{w}_j , \,  \mathbf{o}_{\hat{k}''}]) \enspace \in \mathbb{R}^{2d} \enspace .
\end{equation}

\sloppy We denote our classifier's output for each KE as $f_{KE}\left( \mathbf{e}_{ke_j} \right) = \mathbf{z} = \left(\mathbf{z}_1,\ldots,\mathbf{z}_{|KE|}\right) = \left( f_{KE}\left( \mathbf{e}_{ke_1} \right), \ldots, f_{KE}\left( \mathbf{e}_{ke_{|KE|}} \right) \right)$, where $\mathbf{z}_i$ denotes the output logits of the classifier for $ke_i$ and $f_{KE}(\cdot) \in \mathbb{R}^{3}$.

\subsection{Multi-instance learning losses}
\label{sec:method:mil}
Because we lack KE-level supervision, to train $f_{KE}$ we leverage novel multi-instance learning (MIL) objectives which we derive from the problem semantics.
Specifically, we observe that if a hypothesis is entailed by an image, all KEs within the hypothesis should themselves be entailed (denoted ent.). Formally, $(y = \text{ent}) \implies \forall_{ke_i} (y_{ke_i} =  \text{ent})$. Because there is no ambiguity as to what each KE's label should be, we impose a standard cross entropy classification loss across all KEs for entailed samples:
\begin{equation}
\mathcal{L}_{KE_{y=\text{ent}}} = \sum_{i}^{|KE|} - \log \frac{\exp \left( z_{i_{\text{ent}}} \right)}{\sum_c^C \exp \left( z_{i_c} \right) }
\end{equation}
where $z_{i_{c}}$ is the classifier's predicted value for class $c \in \{\text{ent, neu, con}\}$.

\sloppy For a sample to be labeled neutral, we next observe that its KEs must observe the following definition: $(y = \text{neu}) \implies \left( \forall_{ke_i} \neg \left(y_{ke_i} =  \text{con} \right)  \land \exists_{ke_i} \left(y_{ke_i} = \text{neu}\right) \right)$. That is, no KE may be labeled contradiction (but they may be labeled either entailment or neutral) and at least one KE should be labeled neutral.
We enforce these two constraints through the following MIL loss:
\begin{equation}
\mathcal{L}_{KE_{y=\text{neu}}} = \sum_{i}^{|KE|} \left(- \log \left(1-\frac{\exp \left( z_{i_{\text{con}}} \right)}{\sum_c^C \exp \left( z_{i_c} \right) } \right) \right) - \argmax_{z_{i_{neu}} \in \mathbf{z}} \left(\log \frac{\exp \left( z_{i_{\text{neu}}} \right)}{\sum_c^C \exp \left( z_{i_c} \right) } \right)
\end{equation}
where $\argmax_{z_{i_{neu}} \in \mathbf{z}}$ selects the KE whose score in the neutral dimension is the largest. Intuitively, this amounts to selecting the KE the model is most confident is neutral and treating its label as such.

\begin{figure*}[t]
    \centering
    \includegraphics[width=1\textwidth]{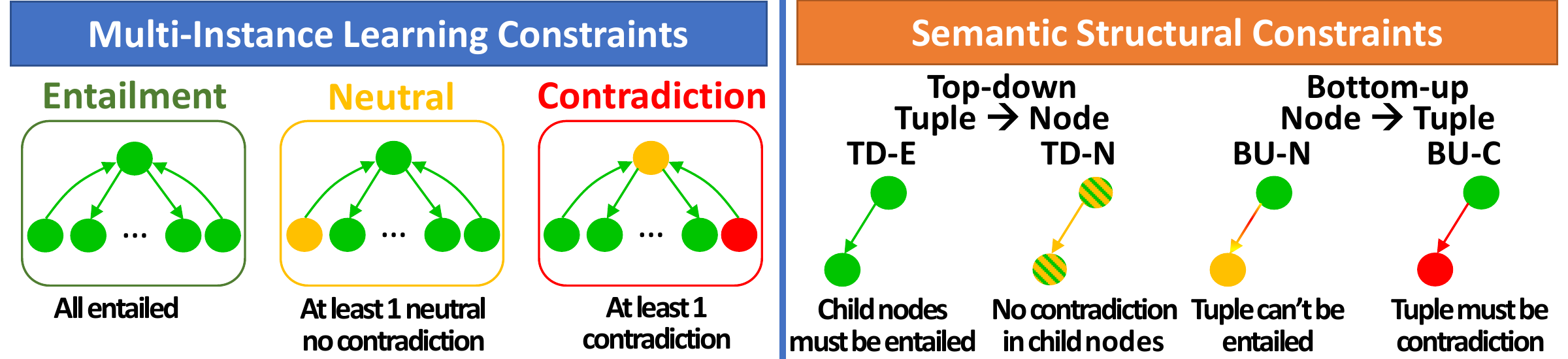}
    \caption{
    \textbf{Left:} Our MIL constraints transfer sample-level supervision to KEs, but allow semantically inconsistent predictions (e.g.~bicycle=contradiction, riding bicycle=entailed). \textbf{Right:} Structural constraints work within the graph to ensure semantically consistent predictions. Multicolor nodes / tuples indicate the KE could be either color.
    }
    \label{fig:method:constraints}
\end{figure*}

Finally, for samples labeled contradiction, we note that $(y = \text{con}) \implies \exists_{ke_i} \left(y_{ke_i} = \text{con}\right)$. In other words, at least one KE must be labeled contradiction. Following the notation used above, we impose the following MIL loss:
\begin{equation}
\mathcal{L}_{KE_{y=\text{con}}} = \argmax_{z_{i_{con}} \in \mathbf{z}} \left(\log \frac{\exp \left( z_{i_{\text{con}}} \right)}{\sum_c^C \exp \left( z_{i_c} \right) } \right)
\end{equation}
which selects the KE the model is most confident is contradiction and enforces it to be so classified. 
We illustrate our MIL constraints for these three categories in Figure \ref{fig:method:constraints} (left).

\subsection{Semantic structural constraints}
The above constraints enforce that the KE predictions are consistent with the sample-level label, but they do not ensure that the KE predictions are \textit{internally} semantically consistent with one another. For example, a model trained with the above constraints would be free to predict the node ``girl'' as contradictory, but the tuple ``girl on bicycle'' as entailed. We call this a ``bottom-up'' semantic structural violation because the tuple's prediction (the parent) is inconsistent with the node's prediction (the child). Like our MIL constraints, our structural constraints flow from the semantics of the problem. We note that two types of bottom-up structural constraints should hold: BU-C) $(y_{ke_i} = \text{con}) \implies \forall_{ke_{j} \in \text{parent}\left(ke_i\right)} \left(y_{ke_j} = \text{con}\right)$ and BU-N) $(y_{ke_i} = \text{neu}) \implies \forall_{ke_{j} \in \text{parent}\left(ke_i\right)} \neg \left(y_{ke_j} = \text{ent}\right)$.
BU-C requires that, if a node is contradiction, any parent tuple that contains it must also be contradiction. BU-N requires that if a node is neutral, no parent tuple may be entailed.  We enforce BU-C and BU-N through the following two bottom-up structure preserving losses:
\begin{equation}
\mathcal{L}_{\text{STRUC}_\text{BU-C}} = \sum_{ke_i, ke_j}^{\mathclap{\substack{ke_i \in V \\ ke_j \in \text{parent}(ke_i)}}} - \sigma \left( z_{i_{\text{con}}} \right) \log \left( \frac{\exp \left( z_{j_{\text{con}}} \right)}{\sum_c^C \exp \left( z_{j_c} \right) } \right) \cdot \mathds{1} \left\{\hat{y_i}=\text{con}\right\} 
\end{equation}
\begin{equation}
\mathcal{L}_{\text{STRUC}_\text{BU-N}} = \sum_{ke_i, ke_j}^{\mathclap{\substack{ke_i \in V \\ ke_j \in \text{parent}(ke_i)}}} - \sigma \left( z_{i_{\text{neu}}} \right) \log \left( 1 - \frac{\exp \left( z_{j_{\text{ent}}} \right)}{\sum_c^C \exp \left( z_{j_c} \right) } \right) \cdot \mathds{1} \left\{\hat{y_i}=\text{neu}\right\}
\end{equation}
where $\sigma$ is the sigmoid function, $\hat{y_i}$ represents $ke_i$'s predicted label (i.e.,~the maximum scoring class in $z_i$), and $\mathds{1}$ is the indicator function. Note that we weight each structural constraint with the confidence of the child's prediction which lessens the impact of incorrectly predicted KEs. We found that this significantly improved performance.

\sloppy Similarly, two top-down constraints must also hold for their predictions to be logically consistent: TD-E) $(y_{ke_i} = \text{ent}) \implies \forall_{ke_{j} \in \text{child} \left(ke_i\right)} \left(y_{ke_j} = \text{ent}\right)$ and TD-N) $(y_{ke_i} = \text{neu}) \implies \forall_{ke_{j} \in \text{child} \left(ke_i\right)} \neg \left(y_{ke_j} = \text{con}\right)$. Analogous to our bottom-up constraints, we enforce our top-down constraints through the following two losses:
\begin{equation}
\mathcal{L}_{\text{STRUC}_\text{TD-E}} = \sum_{ke_i, ke_j}^{\mathclap{\substack{ke_i \in V \\ ke_j \in \text{child}(ke_i)}}} - \sigma \left( z_{i_{\text{ent}}} \right) \log \left( \frac{\exp \left( z_{j_{\text{ent}}} \right)}{\sum_c^C \exp \left( z_{j_c} \right) } \right) \cdot \mathds{1} \left\{\hat{y_i}=\text{ent}\right\} 
\end{equation}
and
\begin{equation}
\mathcal{L}_{\text{STRUC}_\text{TD-N}} = \sum_{ke_i, ke_j}^{\mathclap{\substack{ke_i \in V \\ ke_j \in \text{child}(ke_i)}}} - \sigma \left( z_{i_{\text{neu}}} \right) \log \left(1 -  \frac{\exp \left( z_{j_{\text{con}}} \right)}{\sum_c^C \exp \left( z_{j_c} \right) } \right) \cdot \mathds{1} \left\{\hat{y_i}=\text{neu}\right\}
\, .
\end{equation}

\subsection{Final loss formulation}
In addition to our KE-level losses, we also include a standard sample-level cross-entropy loss performed on the CLS token of the transformer which we denote by $\mathcal{L}_{CLS}$. Thus, our final loss formulation is given by the summation of the previous losses: $\mathcal{L} = \beta_{\text{CLS}} * \mathcal{L}_{\text{CLS}} + \beta_{\text{KE}} * \mathcal{L}_{\text{KE}} + \beta_{\text{STRUC}} * \mathcal{L}_{\text{STRUC}}$, where $\beta$ are hyperparameters controlling the relative weight of each component of the loss. 

\subsection{Implementation details}
All methods and baselines use a pretrained VinVL \cite{zhang2021vinvl} multimodal transformer as our backbone architecture with the ResNeXt-152 C4 detector for visual features. We use a max length of 50 for $\mathbf{o}$ and 165 for $|\mathbf{t}|+|\mathbf{r}|+|\mathbf{h}|$, truncating $\mathbf{h}$. We use a batch size of 128, an initial learning rate of $5e^{-5}$ that linearly decreases, a weight decay of 0.05, and train for a max of 10 epochs. We use Spring \cite{bevilacqua-etal-2021-one} to extract AMR graphs from hypotheses. We use a depth-first approach for AMR linearization. We implement our model in PyTorch \cite{paszke2019pytorch}. Training takes approximately two days on four Nvidia Titan RTX GPUs. Unless otherwise specified, we set $\beta_{\text{CLS}}=0.5$ and $\beta_{\text{KE}}=\beta_{\text{STRUC}}=1$. We include additional details in supplementary.

%% file: results.tex
\section{Experiments}
\label{sec:results}
We compare our method to several baselines on the new task of fine-grained visual entailment. Our results consistently demonstrate that our approach significantly outperforms these baselines on the fine-grained visual entailment task. We also include detailed ablations and analysis of various components of our method. We also present qualitative results illustrating that our method makes semantically meaningful predictions at the KE-level. Finally, we show the visual regions chosen by our method to make its predictions for each KE.

\subsection{Dataset}
The original visual entailment benchmark \cite{xie2019visual} was found to have a substantial ($\sim$39\%) label error rate for the neutral class 
\cite{kayser2021vil}. For training and testing, we therefore use the relabeled version presented in \cite{kayser2021vil} which corrects this issue in the test set. The dataset contains 430,796 image, hypothesis, label triples in total. We use the original train/val/test splits \cite{kayser2021vil}. 

In order to evaluate our method's on the KE-level prediction task, we require KE-level annotated data. To do so, we created an web annotation interface using LabelStudio \cite{labelstudio}. Our interface shows annotators the image, hypothesis, and an image of the hypothesis AMR graph. Annotators annotate each node and tuple (the KEs) within the graph with the class that describes its relationship to the image. Note that we also allow annotators to ``opt-out'' of KEs that are of unclear meaning (which may occur from AMR prediction errors, etc.). These KEs are ignored in evaluation. Because the original sample labels are crowdsourced and still noisy, we also ask annotators to provide a new sample-level label.

Annotating AMR graphs is an intellectually demanding and laborious task, often requiring annotators to consult PropBank \cite{palmer2005proposition} or the AMR specifications \cite{amr3} to understand the meaning of the KE. Because of the difficulty of the task, we concluded it was inappropriate for crowdsourcing. We instead employed two expert  annotators who were familiar with AMR, similar to \cite{banarescu2013abstract, bonial2018abstract}. Collectively, our annotators annotated 1909 KEs (1113-e, 306-n, 282-c, 208-opt-out) from 300 random samples (100-e, 100-n, 100-c) from the test set. Our annotation process took $\sim$75 hours of effort. We include more details in our supplementary.

\subsection{Baselines} 
Because we are the first to tackle the fine-grained visual entailment task, there are no standard baselines for this task. We thus formulate two baselines in order to benchmark our method's performance. 
VE$\rightarrow$KE is a standard sample-level visual entailment model which takes as input the visual features and the hypothesis. We replicate the model's sample-level prediction for every KE.  
VE+AMR$\rightarrow$KE is similar to the previous model but also takes as input the linearized AMR of the hypothesis. At test time, we make a prediction for each KE separately by feeding the AMR corresponding to each KE into the model.

\subsection{Quantitative results}

\input{_tab_ke_level}

We experimentally compare several variants of our method with the baselines described above. Ours is our full method described in Sec.~\ref{sec:method}. We also include an ablation of our method showing the performance of our method without our structural constraints (w/o $\mathcal{L}_\text{STRUC}$) and without our KE-specific region retrieval technique (w/o Region Retrieval). The latter can be directly applied on top of encoders not based on object features as well. Finally, we show a version of our method (Ours+CLS) that predicts all KEs as entailed when the sample-level is predicted entailed. Otherwise, it uses our KE-specific classifier.

\input{_tab_sample_level}
\noindent \textbf{KE-level performance.} In Table \ref{table:results:ke_level} we show the performance of each method on our KE-level annotations. The first group of results shows the accuracy for each ground truth KE class. VE$\to$KE achieves strong performance for KEs labeled neutral and particularly contradiction. This is because contradiction KEs only appear in contradiction samples, while neutral and entailment KEs can appear in multiple categories.  However, we observed that relatively few neutral KEs appear in the contradiction category, because hypotheses usually mention untrue facts (contradiction) about something truly in the image (entailment). Thus, performance is also strong for the neutral category, since neutral KEs largely only appear in neutral samples. We note that the KE classes are unbalanced by nature of our task and we seek to locate the entailment KEs in neutral and contradiction samples. To this end, our method substantially outperforms both baselines for entailment KEs which can appear in any type of sample and are thus by far the most frequent. This performance is reflected in the node, tuple, and overall accuracies on which our method performs best. 

The second group of results shows accuracies across different types of KEs. We note our method performs better on nodes than tuples. This is unsurprising because nodes represent simpler statements about the presence of objects or actions, while tuples make more complex, composite statements about nodes and are thus harder to verify for the model. While VE+AMR$\to$KE outperforms VE$\to$KE overall and for most metrics, their predictions are highly similar for most samples and it is significantly outperformed by our method overall.

We next measure the number of structural violations. We consider each parent-child KE relationship a separate instance and calculate the accuracy of each method at producing semantically consistent predictions across parent-child relationships. We note that while VE$\to$KE has no structural violations (because all KEs are predicted as the sample label), it makes no fine-grained predictions. Of the methods that make such predictions, our method performs best.

Finally, we measure the performance of our ablated method and our method's variants. Without $\mathcal{L}_\text{STRUC}$ our method achieves the lowest overall performance of our method, indicating that the structural constraints work synergistically with our MIL constraints to further disambiguate the KE-level labels. We further note that without $\mathcal{L}_\text{STRUC}$ our model has a high rate ($\sim 30\%$) of structural violations within the graph. We next show that aggregating visual features into our model's contextual embedding is important. Without region retrieval our model's accuracy drops by (-3.45\% acc) because our KE's embeddings lack relevant visual context (especially for nodes). Finally, we observe that ignoring our KE-level predictions for sample's predicted entailment and predicting all KEs as entailed (Ours+CLS) slightly improves performance (+0.11\% acc).

\begin{figure*}[t]
    \centering
    \includegraphics[width=1\textwidth]{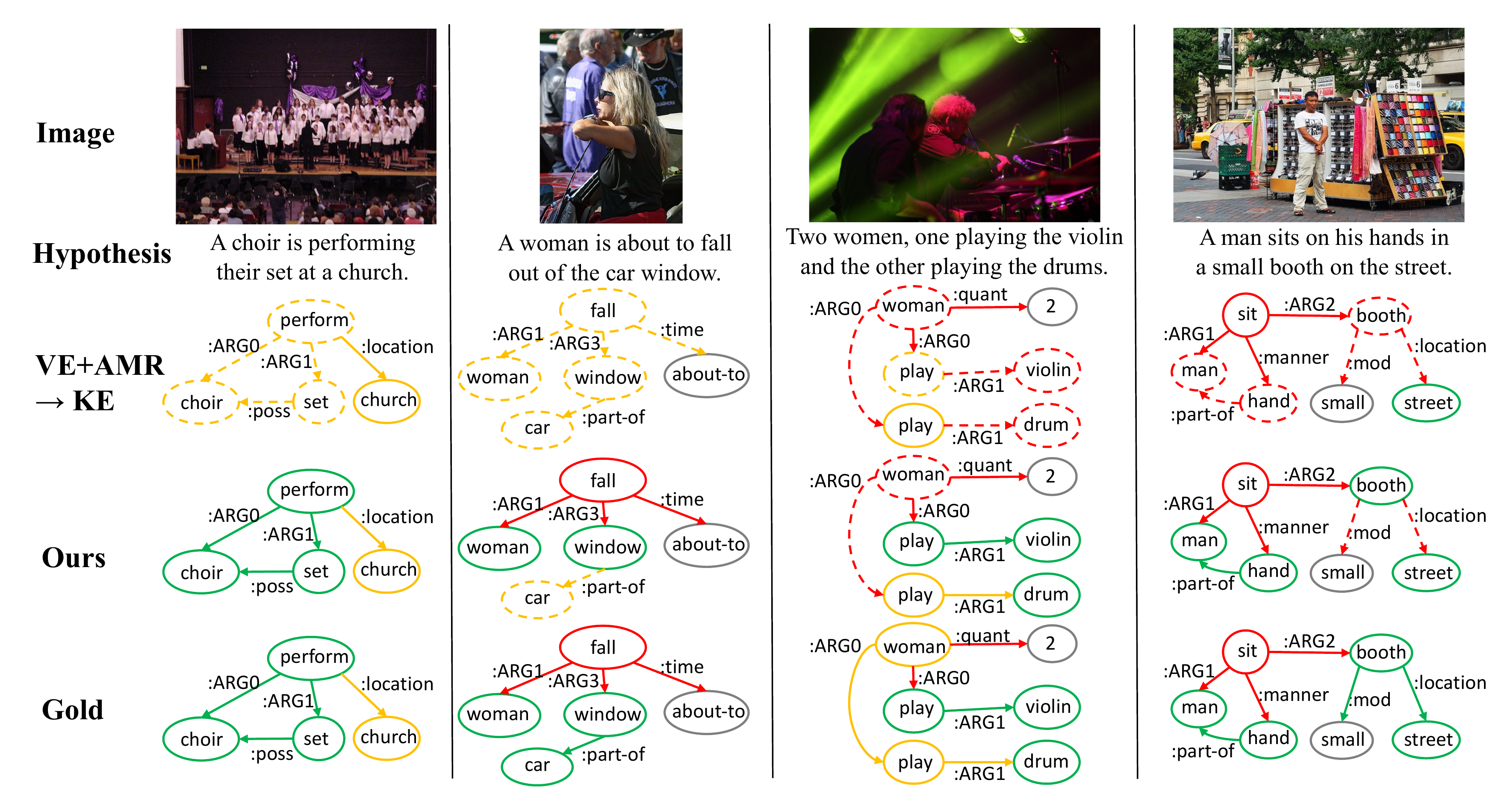}
    \vspace{-2em}
    \caption{
    Qualitative results showing our KE-level predictions on AMR graphs compared to the baseline. Nodes and edges (representing tuples) are colored based on their predicted label ({\color{Green} ent}, {\color{Yellow} neu}, {\color{Red} con}, {\color{Gray} opt-out}). Wrong predictions are denoted by dashed lines. %
    }
    \label{fig:results:qual}
    \vspace{-2em}
\end{figure*}

\noindent \textbf{Sample-level performance.} Though it is not our focus, we also include the performance of different methods at predicting the sample-level label in Table \ref{tab:results:sample_level}. The left side shows the performance on the labels given by the MTurkers in \cite{kayser2021vil}, while the right shows the performance on the sample labels in our expertly annotated set (Relab.). We explore two ways of predicting the sample label. The first uses the CLS token, while the second uses the logical rules defined in Sec.~\ref{sec:method:mil} to produce the sample label from the predicted KE labels. We observe a slight drop (0.38\%) in the sample-level label for our method on the crowdsourced labels. Aside from label noise, one possible reason is that the model pays less attention to the sample-level task and focuses on the KE-level task (see ablation on loss weightings in our supplementary for more details). We show that on our set of expertly annotated samples, producing the sample-level label using our KE predictions outperforms all baselines for sample-level prediction.

\subsection{Qualitative results}
In this section, we present qualitative results showcasing our method's KE-level predictions. We also illustrate the visual region our method selects for each KE.

\noindent \textbf{KE-level prediction results.} In Figure \ref{fig:results:qual}, we show KE-level prediction results.
We observe that the baseline often predicts many KEs the same label. In contrast, our model's predictions are more diverse and reasonable. In the first column, our model correctly concludes the location can't be determined and marks ``\texttt{church}'' neutral. In the next column, our model struggles to detect a car in the image so incorrectly marks ``\texttt{car}'' as neutral, but correctly infers that ``\texttt{fall}'' is contradictory. %
In the rightmost column, our model ``incorrectly'' says the booth is not small, but this is subjective. Similarly, our model says the booth is not located on the street,
which is feasible because the booth is on the sidewalk.

\begin{figure*}[t]
    \centering
    \includegraphics[width=1\textwidth]{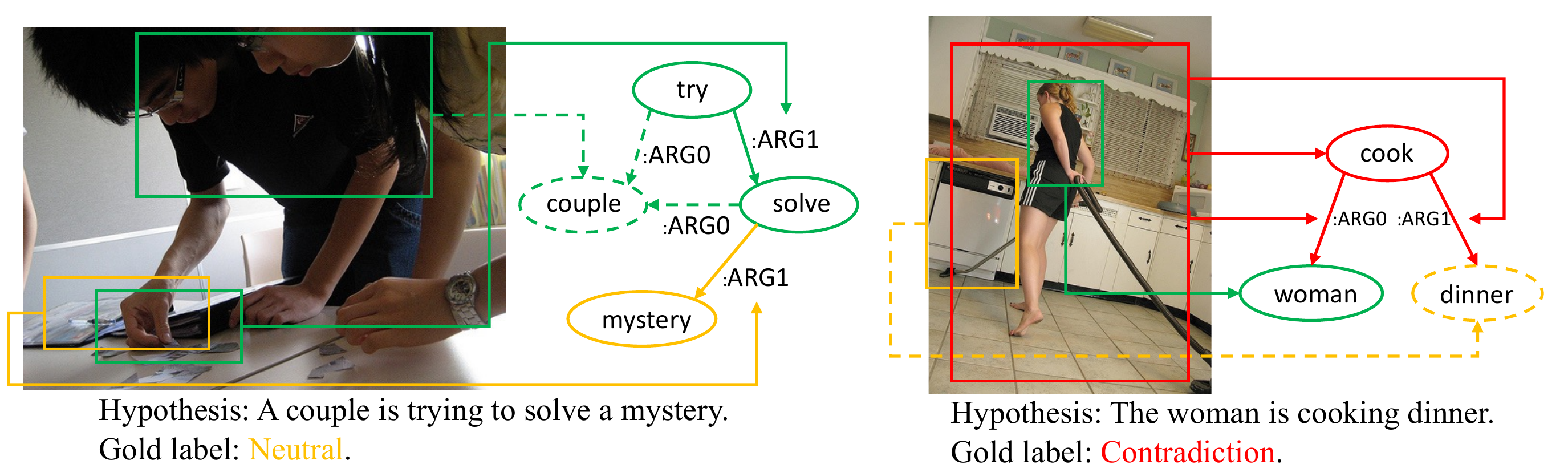}
    \caption{
    We show the image region our model selects for each KE prediction. KEs are colored based on their predicted label. Wrong predictions are denoted by dashed lines.
    }
    \label{fig:results:attn}
    \vspace{-1.5em}
\end{figure*}

\noindent \textbf{Region selection results.} In Figure \ref{fig:results:attn} we show the 
image regions selected by our model for each KE. We observe relevant KE$\to$image localization results. For example, on the left we observe that ``\texttt{(try, :ARG1, solve)}'' retrieves the puzzle being assembled and ``\texttt{couple}'' retrieves the two people. Nevertheless, the ground truth label for ``\texttt{couple}'' is neutral. Our model could be deceived by the proximity of the two people in the retrieved region. On the right, the model selects the upper body of the woman for ``\texttt{woman}'' and a large region of the image to conclude the action ``\texttt{cook}'' is incorrect.

%% file: _tab_ke_level.tex
\setlength{\tabcolsep}{4pt}
\begin{table}[t]
\begin{center}
\caption{We show KE-level accuracies at the class-level, across different types of KEs, and the overall KE-level accuracy. Finally, we show the structural constraint accuracy (see text). The best result per column is shown in bold and second best is underlined. }
\label{table:results:ke_level}
\vspace{-1em}
\resizebox{1\linewidth}{!}{
\begin{tabular}{l|ccc|cc|cc}
\hline\noalign{\smallskip}
Method & Acc$_{\text{ent}}$ & Acc$_{\text{neu}}$ & Acc$_{\text{con}}$ & Acc$_\text{node}$ & Acc$_\text{tup}$ & Overall Acc & Acc$_\text{STRUC}$ \\
\noalign{\smallskip}
\hline \hline
\noalign{\smallskip}
VE $\to$ KE & 49.77 & \textbf{60.00} &  \underline{83.33} & 54.36 & 60.51 & 57.17 & \textbf{100} \\
VE+AMR $\to$ KE & 55.51 & \underline{58.06} & \textbf{85.10} & 58.14 & 64.10 & 60.86 & 96.40\\
\noalign{\smallskip}
\hline
\noalign{\smallskip}
w/o $\mathcal{L}_{\text{STRUC}}$ & \textbf{88.87} & 15.48 & 9.57 & 66.88 & 57.17 & 62.44 & 70.26\\
w/o Region Retrieval & 78.20 & 24.51 & 55.67 & 64.61 & 64.87 & 64.73 & 93.57\\
Ours  & 79.64 & 31.29 & 62.76 & \underline{69.79} & \underline{66.02} & \underline{68.07} & 96.36\\
Ours+CLS  & \underline{80.35} & 29.35 & 62.76 & \textbf{70.01} & \textbf{66.02} & \textbf{68.18} & \underline{96.98} \\
\hline
\end{tabular}
}
\end{center}
\vspace{-2em}
\end{table}
\setlength{\tabcolsep}{1.4pt}

%% file: _tab_sample_level.tex
\setlength{\tabcolsep}{4pt}
\begin{table}[t]
\begin{center}
\caption{We show the sample-level accuracy of each method across different metrics. The best method for each metric is shown in bold and the second best is underlined.}
\label{tab:results:sample_level}
\begin{tabular}{l|cc|ccc}
\hline\noalign{\smallskip}
Method & $\text{Acc}_{\text{CLS}}$ & $\text{Acc}_{\text{KE}\to\text{CLS}}$ & $\text{Acc}_{\text{CLS}}^{\text{Relab.}}$ & $\text{Acc}_{\text{KE}\to\text{CLS}}^{\text{Relab.}}$ & $\text{Acc}_{\text{Best}}^{\text{Relab.}}$\\
\noalign{\smallskip}
\hline \hline
\noalign{\smallskip}
VE $\to$ KE & \textbf{80.37} & - &  \textbf{79.73} & - & 79.73\\
VE+AMR $\to$ KE & - & \textbf{79.15} & - & \underline{80.06} & \underline{80.06}\\
\noalign{\smallskip}
\hline
\noalign{\smallskip}
w/o $\mathcal{L}_{\text{STRUC}}$ & 79.78 & 75.17 & \underline{79.40} & 79.73 & 79.73\\
w/o Region Retrieval & 79.58 & 75.21 & \underline{79.40} & \underline{80.06} & \underline{80.06} \\
Ours  & \underline{79.99} & 76.26 & 78.73 & \textbf{80.73} & \textbf{80.73} \\
Ours+CLS  & \underline{79.99} & \underline{77.70} & 78.73 & 77.74 & 78.73 \\
\hline
\end{tabular}
\end{center}
\vspace{-2em}
\end{table}
\setlength{\tabcolsep}{1.4pt}

%% file: conclusions.tex
\section{Conclusion}
\vspace{-0.5em}
We introduced the novel problem of fine-grained visual entailment where the goal is to predict the logical relationship of knowledge elements extracted from a piece of text with an image. We proposed a model for this task which fuses relevant visual features with the representation of each knowledge element.  Because of a lack of fine-grained annotations, we proposed novel multi-instance learning losses to transfer sample-level supervision to the knowledge element-level. We also proposed novel semantic structure preserving constraints. Experiments conducted on a new benchmark show that our approach significantly outperforms relevant baselines, and more importantly, produces interpretable predictions. 

There are several possible directions for future work. For example, pretraining encoders with human-created AMR inputs \cite{hinton2015distilling,bevilacqua-etal-2021-one} may better prepare encoders for our task. While we use object labels in our work, object attributes are also potentially helpful for predicting KEs that involves attributes \cite{zhang2021vinvl}. Finally, region retrieval \cite{kumar2020region, babar2021look, yuan2020weakly} and graph networks \cite{velivckovic2018graph} may also be exploited.

%% file: _tab_ablation.tex
\setlength{\tabcolsep}{4pt}
\begin{table}[th]
\begin{center}
\caption{Ablation of different loss weights used for training our model (Ours). For the results in this table, only the KE classifier $f_{\text{KE}}$ is used. The best result per column is shown in bold and second best is underlined.
}
\label{table:supp:ablation}
\resizebox{1\linewidth}{!}{
\begin{tabular}{ccc|ccc|cc|ccc}
\hline\noalign{\smallskip}
$\beta_{\text{CLS}}$ & $\beta_{\text{KE}}$ & $\beta_{\text{STRUC}}$ & Acc$_{\text{ent}}$ & Acc$_{\text{neu}}$ & Acc$_{\text{con}}$ & Acc$_\text{node}$ & Acc$_\text{tup}$ & $\text{Acc}_{\text{KE}\to\text{CLS}}^{\text{Relab.}}$ & Acc$_\text{KE}$ & Acc$_\text{STRUC}$ \\
\noalign{\smallskip}
\hline \hline
\noalign{\smallskip}
1 & 1 & 1 & 74.61 & 27.09 & 30.85 & 62.45 & 54.35 & 67.77 & 58.75 & 94.80 \\
0.1 & 1 & 1 & \underline{84.75} & \underline{30.64} & \underline{44.68} & \textbf{71.84} & \underline{64.10} & 75.41 & \textbf{68.30} & \underline{96.46} \\
0.5 & 1 & 1 & 79.64 & \textbf{31.29} & \textbf{62.76} & \underline{69.79} & \textbf{66.02} & \textbf{80.73} & \underline{68.07} & 96.36 \\
0.5 & 0.5 & 1 & 76.50 &	\underline{30.64} & 37.23 & 64.29 &	58.58 & 61.79 & 61.68 & \textbf{98.40} \\
0.5 & 1 & 0 & \textbf{88.87} & 15.48 & 9.57 & 66.88 & 57.17 & \underline{79.73} & 62.44 & 70.26 \\
\hline
\end{tabular}
}
\end{center}
\end{table}
\setlength{\tabcolsep}{1.4pt}